\documentclass{article}

\usepackage[preprint,nonatbib]{neurips_2026}

\usepackage[utf8]{inputenc} 
\usepackage[T1]{fontenc}    
\usepackage{hyperref}       
\usepackage{url}            
\usepackage{booktabs}       
\usepackage{amsfonts}       
\usepackage{nicefrac}       
\usepackage{microtype}      
\usepackage{xcolor}         
\usepackage{graphicx} 
\usepackage[numbers]{natbib}
\usepackage{multirow}
\usepackage{multicol}
\usepackage{amsmath}
\usepackage{amsfonts}
\usepackage{geometry} 

\title{SUMI: Scalable Unified Model for 3D Point Cloud Inference}

\author{%
  Yanlong LI\thanks{Use footnote for providing further information
    about author (webpage, alternative address)---\emph{not} for acknowledging
    funding agencies.} \\
  School of Computer Science\\
  The University of Sydney\\
  Sydney, NSW 2006 \\
  \texttt{yali8838@uni.sydney.edu.au} \\
  \And
  Kanchana Thilakarathna \\
  School of Computer Science\\
  The University of Sydney\\
  Sydney, NSW 2006 \\
  \texttt{kanchana.thilakarathna@sydney.edu.au} \\
}

\begin{document}

\maketitle
\newcommand{\draft}[1]{{\color{orange}#1}}
\newcommand{\keynote}[1]{{\color{red}#1}}
\begin{abstract}
Point cloud completion commonly follows a coarse-to-fine paradigm, where a low-density coarse shape is first predicted and then upsampled to the target resolution. Although recent methods have improved global structure recovery, the fine stage often remains limited by simple upsampling and insufficient interaction with coarse structural features, making local detail reconstruction challenging. We propose SUMI, a diffusion-enhanced refinement module for coarse-to-fine point cloud completion. Unlike prior diffusion-based completion methods that use diffusion as a standalone point generator, SUMI injects noisy geometric features into cross-attention with coarse structural features, enabling reverse denoising to refine local geometry while preserving global consistency. SUMI can also be integrated into existing coarse-to-fine models as a flexible refinement module. Experiments on PCN, ShapeNet-55/34, and MVP demonstrate consistent improvements over strong baselines. SUMI achieves the best overall CD and F1-score on PCN, reduces CD by up to 16.1\% on ShapeNet-55, and obtains the best CD across all output densities on MVP.
\end{abstract}

\section{Introduction}

With the rapid development of 3D sensing and reconstruction technologies, point clouds have become a fundamental representation for modeling real-world geometry due to their flexibility, efficiency, and direct compatibility with modern 3D capturing technologies such as LiDAR, RGB-D cameras, and multi-view reconstruction systems. They are widely used in downstream tasks including autonomous driving, robotics, augmented and virtual reality (AR/VR), digital twins, and 3D content creation. However, in practical applications, due to factors such as limited sensor field of view, occlusion, insufficient sampling density, and noise interference, the acquired point cloud data often suffer from varying degrees of missing and incomplete information. Incomplete geometric information can significantly affect the performance of downstream tasks, such as 3D recognition, segmentation, and reconstruction. Therefore, the point cloud completion problem has emerged, with the goal of recovering a complete, dense, and geometrically consistent 3D shape from a partial point cloud.

Early point cloud completion methods mainly relied on traditional techniques such as geometric priors, rule constraints, or template matching \cite{kazhdan2006poisson,li2015database,pauly2005example}. These methods typically assume that objects have specific structural patterns or topological priors, and can achieve certain results in simple scenes. However, their generalization ability and robustness are significantly limited when faced with complex object shapes or cross-class scenes. With the development of deep learning, researchers have begun to utilize data-driven methods to learn implicit or explicit shape priors, significantly advancing the field of point cloud completion \cite{han2017high,yu2021pointr,wang2021voxel,wang2018adaptive}. These methods typically map a portion of the point cloud to a latent representation space using an encoder, and then generate the complete point cloud using a decoder, achieving significant improvements in completion accuracy and stability. Recent advances in representation learning have introduced more expressive architectures, such as attention mechanisms, masked modeling, and generative frameworks, to better capture long-range dependencies and global shape priors. By learning from partially observed data and exploiting large-scale pretraining or generative sampling, these approaches show improved robustness and diversity in completion results. 

Many 3D objects exhibit geometric symmetry, making symmetry priors useful for point cloud completion. While early symmetry-based methods relied mainly on global transformations, they often struggled to recover accurate local geometric relationships. Recent work~\cite{yan2025symmcompletion} addresses this by introducing LSTNet, which leverages local symmetry information to improve missing-region reconstruction. Nevertheless, most deep learning-based methods still adopt a coarse-to-fine pipeline, where a low-density coarse output is further upsampled to the target resolution. This makes final completion quality dependent on the upsampling stage and the limited information carried by the coarse representation.

To address the aforementioned issues, we propose SUMI, a diffusion-enhanced refinement mechanism within the coarse-to-fine framework for high-quality point cloud completion. Different from prior diffusion-based completion methods that directly generate point coordinates through a standalone denoising process, SUMI uses diffusion noise as a fine-stage refinement signal inside feature interaction. By injecting noisy geometric features into cross-attention with the coarse representation, SUMI progressively enhances local geometric details while preserving the global structure provided by the coarse prediction. Therefore, SUMI is not designed to replace the coarse-to-fine pipeline, but to strengthen its fine-generation stage and can be integrated into existing coarse-to-fine completion models as a flexible refinement module. Our main contributions are:

\begin{itemize}
    \item We propose SUMI, a diffusion-enhanced fine-stage refinement module that embeds denoising into feature interaction rather than directly generating point coordinates.
    \item We introduce a noise-conditioned cross-attention mechanism that injects noisy geometric features into coarse structural representations for local detail recovery.
    \item We show that SUMI can be integrated into existing coarse-to-fine methods without major architectural changes, demonstrating its flexibility as a refinement module.
    \item We evaluate the SUMI module through targeted ablation studies, analyzing the effects of output density, insertion stage, and diffusion timestep.
\end{itemize}

\begin{figure}
  \centering
  \includegraphics[width=0.95\linewidth]{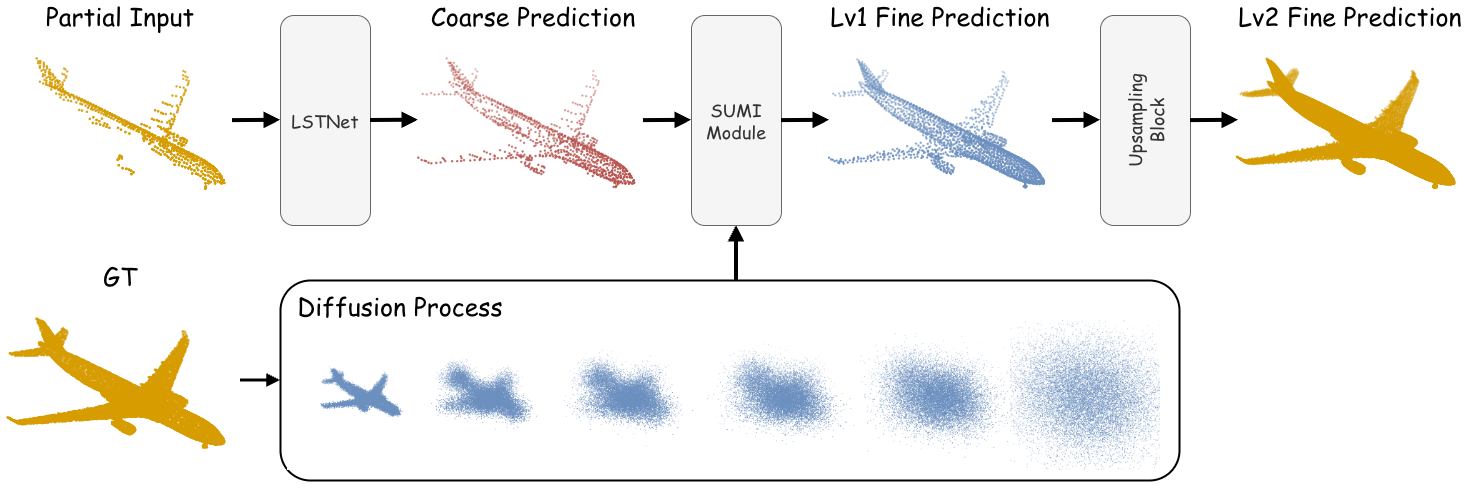}
  \caption{Overview of SUMI. A coarse prediction is first generated from the partial input, followed by diffusion-enhanced Level-1 refinement and lightweight upsampling for the final output.}
  \label{fig:pipeline}
\end{figure}


\section{Related Work}

\begin{figure}
  \centering
  \includegraphics[width=1\linewidth]{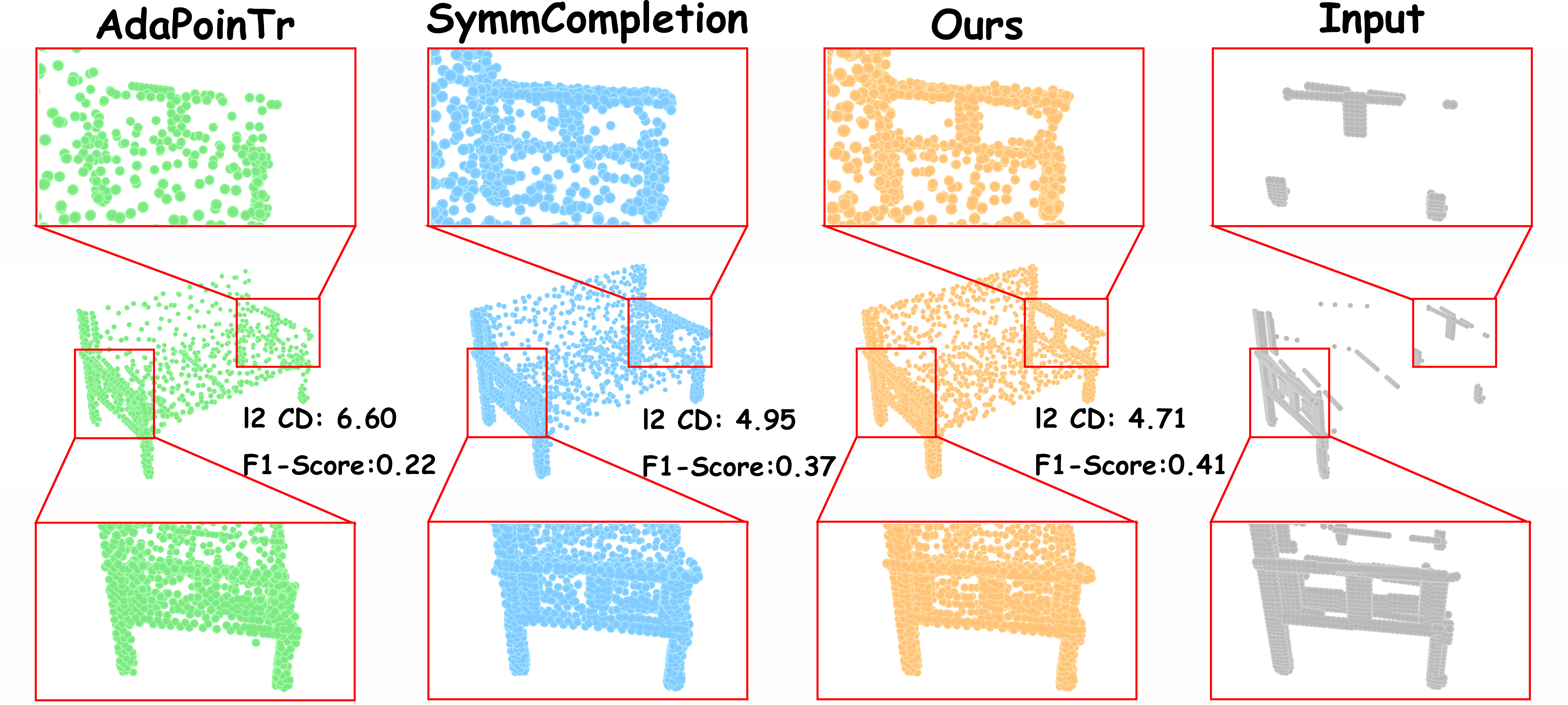}
    \caption{Level-1 fine prediction comparison. From left to right: AdaPoinTr, SymmCompletion, Ours, and the partial input. SUMI reconstructs better local details and more uniform point distributions. Per-sample $L_2$ Chamfer Distance $\times 10^4$ (CD, $\downarrow$) and F1-Score@1\% (F1, $\uparrow$) are shown.}
  \label{fig:lv1_fine}
\end{figure}

\subsection{Point Cloud Completion (PCC)}

Early PCC approaches primarily relied on geometric priors, template matching, or symmetry assumptions to recover missing regions \cite{kazhdan2006poisson,yang2017shape,li2015database,pauly2005example,kazhdan2004symmetry,mitra2006partial}, which limited their applicability to simple shapes and constrained scenarios. With the advancement of deep learning, PCC methods have evolved into two different types. The first type focuses on predicting globally completed point clouds directly from partial input, typically by encoding the incomplete observation into a latent representation that captures overall shape structure, followed by decoding to generate a complete point set \cite{yang2018foldingnet,fu2023vapcnet,wang2024pointattn}. These methods benefit from strong global shape reasoning but may suffer from loss of fine-grained geometric details. The second type directly focuses on inferring missing regions conditioned on visible points, explicitly modeling the relationship between observed and unobserved geometry \cite{yu2021pointr,zhou2022seedformer}. 

\subsection{Transformer}
Transformers, originally introduced for NLP \cite{vaswani2017attention}, leverage self-attention to model long-range dependencies and global context, achieving strong performance in language understanding and generation \cite{devlin2019bertpretrainingdeepbidirectional,openai2024gpt4technicalreport}. Their success has driven adoption in computer vision, where ViT \cite{dosovitskiy2020image} represents images as token sequences and has shown competitive results across classification, detection, and segmentation tasks \cite{touvron2021trainingdataefficientimagetransformers,he2022masked,manzari2023medvit,li2023uniformer,zong2023detrs,zhao2024detrs,lv2024rtdetrv2improvedbaselinebagoffreebies,liu2025tfnet,kumar2024grey,hu2024patrans,liu2025cswin}. Recently, Transformers have been extended to 3D tasks, including point cloud classification, segmentation, and generation, demonstrating strong global reasoning and robustness \cite{long2023pointclustering,wu2025spiking,Liu_2025_CVPR,kolodiazhnyi2024oneformer3d,xiao2025position,li2025point,lu20253dlst,wu2024text2lidar,lv2024sgformer}. In point cloud completion, PoinTr \cite{yu2021pointr} and subsequent works \cite{zhou2022seedformer,wen2022pmp,li2025dapointr} establish the effectiveness of Transformer-based architectures.

\subsection{Diffusion model (DM)}
DMs have emerged as powerful generative frameworks that iteratively add and remove noise, enabling stable training and high-quality synthesis \cite{sohl2015deep,ho2020denoising}. Initially developed for image generation \cite{croitoru2023diffusion}, they have achieved strong performance across tasks such as image generation, super-resolution, and inpainting \cite{batzolis2021conditional,Yu_2023_ICCV,Zhu_2023_CVPR,li2024return,gao2023implicit,shang2024resdiff,xiao2023ediffsr,niu2024acdmsr,wang2025inpdiffusion,kim2025rad}. Recent work extends DMs to 3D tasks, including point cloud, shape, and scene generation \cite{romanelis2025efficient,luo2021diffusion,wu2023sketch,vahdat2022lion,xiong2025octfusion,zhang20233dshape2vecset,zhou20213d,liu2024reconx,wang2025vistadream,ju2024diffindscene}. By modeling complex geometric distributions, DMs are well suited for irregular point clouds and enable progressive refinement of structure and detail. In PCC, diffusion-based methods were first introduced by PVD \cite{zhou20213d}, followed by subsequent extensions \cite{kasten2023point,du2025superpc}. PDR \cite{lyu2021conditional} further explores conditional diffusion refinement for point cloud completion. In contrast, SUMI embeds noisy geometric information into feature-level cross-attention and uses diffusion as a fine-stage refinement signal within a coarse-to-fine framework.

\section{Methodology}

\begin{figure}
  \centering
  \includegraphics[width=1\linewidth]{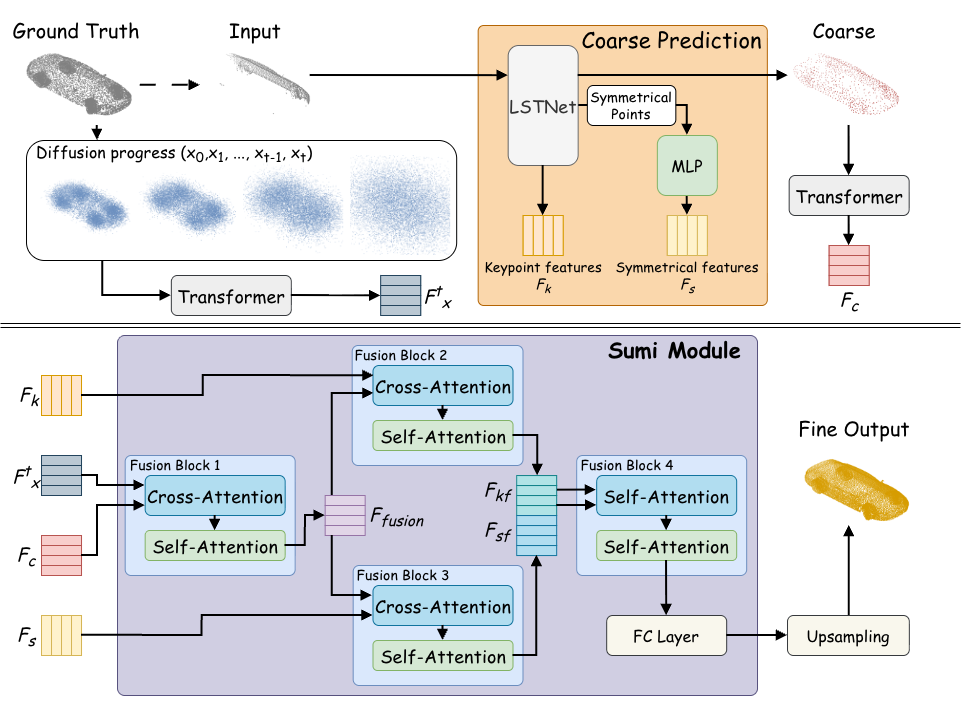}
  \caption{Detailed architecture of the SUMI module. The coarse prediction and noisy point cloud are encoded and fused through noise-conditioned cross-attention, followed by keypoint and symmetry feature fusion for fine prediction.}
  \label{fig:structure}
\end{figure}

\subsection{Overview}


Fig.~\ref{fig:pipeline} presents the overview of our three-layer coarse-to-fine pipeline, which consists of one coarse prediction layer and two fine-generation layers. The coarse layer first predicts a low-density complete shape from the partial input and extracts structural features from the input and its symmetric prediction. During training, Gaussian noise is progressively added to the ground truth, and the noisy point cloud is fused with the coarse representation through Transformer-based feature interaction. This enables SUMI to learn a noise-conditioned refinement process that recovers fine geometric details while preserving global structure. During inference, SUMI performs reverse denoising over $T$ steps to progressively generate the final completed point cloud.

\subsection{Coarse Generation}
For the coarse generation, we take the standard LSTNet module \cite{yan2025symmcompletion} to make the coarse prediction. This module is based on a local point-wise symmetry transformation that maps the existing geometric structure in the partial point cloud to the missing region, thereby constructing an initial point cloud representation with high consistency and fidelity. 

\subsection{SUMI Module}
We apply the SUMI Module for Level-1 Fine generation. We demonstrate the performance of our proposed method for the Level-1 Fine generation and comparison with prior methods in Fig. \ref{fig:lv1_fine}.

\subsubsection{Diffusion Process}
SUMI is based on the standard diffusion framework. We apply diffusion noise in the point-coordinate space before feature encoding, allowing the model to learn recovery from noisy geometry through attention-based feature interaction. During the forward diffusion process, Gaussian noise is recursively added to the ground truth $x_0$, generating a series of intermediate states $x_1, x_2, ..., x_t$ over $T$ steps until the distribution approaches Gaussian noise. Following DDPM and prior works, we model this process as a Markov process as in Eq. \ref{eq:markov}:

\begin{equation}
  q(x_t|x_{t-1}) = \mathcal{N}(x_t;\sqrt{1-\beta_t}x_{t-1},\beta_t\mathbf{I})
  \label{eq:markov}
\end{equation}

During training, we first fuse the coarse features and noise to enhance the model's ability to model complex distributions. Then, keypoint features $F_k$ and symmetry features $F_s$ are used as conditional information, along with the fused features, to predict the clean fine-level point cloud from the noisy observation. Through iterative optimisation, the model gradually recovers the target point cloud distribution from the noise. 


In our implementation, the model is trained to predict the clean point cloud $x_0$ from the noisy observation $x_t$. Specifically, given $x_t$ and conditioning features C, the network outputs $x_f$, and we minimize the reconstruction error between $x_f$ and $x_0$: 

\begin{equation}
    \begin{aligned}
        \mathcal{L}_{\text{diff}} = \mathcal{L}_{CD}(x_f, x_0)
    \end{aligned}
  \label{eq:condition}
\end{equation}

This objective is integrated into the overall end-to-end training described in Section 3.6.

\subsubsection{Fusion Block}
In the SUMI Module, we include four Fusion Blocks for aggregating and enhancing features across multiple sources. We follow previous work \cite{hong2023lrm, yan2025symmcompletion}. Each Fusion Block contains a Cross-Attention mechanism to aggregate features and a Self-Attention mechanism to enhance feature representation.

The overall pipeline of feature fusion inside the SUMI Module is shown in Fig.~\ref{fig:structure}. 
The module takes as input the coarse prediction $P_c$, keypoint features $F_k$, symmetry features $F_s$, and the noisy point cloud at time step $t$, $x_t$. 

We first encode $P_c$ and $x_t$ using two Transformer encoders to obtain feature representations $F_c$ and $F_x^t$, respectively. 
These features are then fused through a cross-attention mechanism. Specifically, $F_c$ is used as the Query, while $F_x^t$ is used as the Key and Value, allowing the model to refine coarse structural features using noise-aware geometric information. That process is defined as below Eq. \ref{eq:noise_fusion} and Eq. \ref{eq:cross_attion}:


\begin{equation}
  F_{fusion} = \mathcal{S}(\mathcal{C}(F_c, F^t_x))
  \label{eq:noise_fusion}
\end{equation}
\begin{equation}
  \mathcal{C}(F_c, F^t_x) = CrossAttention(Q=F_c, K=F^t_x, V=F^t_x)
  \label{eq:cross_attion}
\end{equation}

where $\mathcal{S}$ refers to Self-Attention. Since the key and value are derived from noisy inputs $x_t$, the attention weights become implicitly noise-conditioned, allowing stochastic perturbations to influence feature interaction.

Next, we use two separate Fusion Blocks to incorporate keypoint and symmetry features. Specifically, $F_{fusion}$ is fused with $F_k$ to obtain $F_{kf}$, while another branch fuses $F_{fusion}$ with $F_s$ to obtain $F_{sf}$. The two resulting features are then concatenated and fed into the final Fusion Block, which uses self-attention to further enhance the overall feature representation.
We define this process as Eq. \ref{eq:all_fusion}:

\begin{equation}
  F = \mathcal{S}(\mathcal{C}(F_{fusion}, F_k)) \cdot \mathcal{S}(\mathcal{C}(F_{fusion}, F_s))
  \label{eq:all_fusion}
\end{equation}

where $\cdot$ denotes the concatenation operation.

Finally, a fully connected layer maps the fused features to the point cloud space and upsamples them to the target density to generate a high-resolution output $P_f$.

During inference, SUMI starts from a noisy point cloud and progressively refines it through the reverse denoising process. In practice, the predicted clean point cloud is used as the denoised estimate at each step, following the DDPM formulation summarized in the supplementary material. At each timestep, the predicted fine point cloud is used to guide the transition from $x_t$ to $x_{t-1}$, conditioned on the coarse output $P_c$, keypoint features $F_k$, and symmetry features $F_s$. This process is repeated for $T$ steps to obtain the final fine completion result.

\subsection{Implementation setup}
We apply the SUMI Module at the Level-1 fine-generation stage to achieve high-quality detail restoration. Its output density is determined by hyperparameters. For example, in the PCN dataset, when the target point cloud density is 16384, the input density of the SUMI Module is set to 2048 and then further improved to the target resolution using a lightweight upsampling module.

We placed the SUMI Module only in the first-stage fine generation, primarily based on the following considerations: First, the DM itself has high computational overhead; reusing it across multiple fine stages would significantly increase training and inference time and may lead to training instability. Second, if the SUMI Module is used only in the second fine stage, while a simple upsampling method is used in the first fine stage, the initial refinement result will still retain errors from the coarse stage, which will be further amplified in subsequent processes, thereby affecting the final completion quality. Therefore, this design achieves a good balance between performance and efficiency.

\subsection{Enhancement for existing methods}
Based on the design of noise embedding in our SUMI Module, the forward and reverse diffusion processes can be integrated into coarse-to-fine completion methods with minimal architectural modifications to enhance feature representation and completion performance. We integrate SUMI into AdaPoinTr and evaluate the performance in Sec. \ref{sec:pcn_set}. We further validate the performance of injecting the diffusion process into different modules of the existing coarse-to-fine structured model in Sec. \ref{sec:abl}. These results suggest that SUMI can serve as a flexible refinement module for coarse-to-fine completion architectures.

\subsection{Learning Target and Loss Function}
Our proposed method employs a multi-module end-to-end training strategy, using the ground truth of the complete point cloud as the unified learning objective. During training, the results generated at each stage are co-optimised to improve the overall completion quality.

In the loss function design, we use Chamfer Distance as the primary metric to evaluate geometric consistency between the predicted and ground-truth point clouds. Considering the differences in density between outputs at different stages, we further construct a hierarchical composite loss function to jointly constrain the coarse output and results of different precision levels in the fine stage, thereby achieving collaborative optimisation across the entire process. The overall loss function form is shown below:

\begin{equation}
    \begin{aligned}
        \mathcal{L} = \mathcal{L}_{CD}(P_c, x_0) + \mathcal{L}_{CD}(x_f, x_0) + \mathcal{L}_{CD}(P''_f, x_0)
    \end{aligned}
  \label{eq:loss}
\end{equation}

where $x_0$ is the ground truth, $P_c$ represents the prediction result in the coarse stage, $x_f$ is the output generated by the SUMI Module at time step $t$, and $P''_f$ represents the final high-density output from the upsampling module. This multi-scale supervision mechanism can simultaneously constrain the global structure and local details, thereby effectively improving the model's performance.
\section{Experiments}
In this section, we first introduce the dataset and benchmark for the PCC task. Then, we present the results of our model and compare them with several baselines. We also include ablation studies and visual analysis for our model. We train and validate our method on a single NVIDIA RTX4090.

\subsection{Experimental Setup and Evaluation Metrics}
We use PCN \cite{yuan2018pcn} as the primary dataset for training and evaluation, as it provides partial-complete point cloud pairs generated via virtual depth scans, better reflecting real-world LiDAR/RGB-D observations. We further evaluate on ShapeNet-55/34 \cite{wu20153d} and MVP \cite{pan2021mvp} to assess performance on diverse categories and single-view inputs. Unlike prior diffusion methods that often operate on limited ShapeNet subsets \cite{vahdat2022lion, wei2025sc, kasten2023point, du2025superpc}, our evaluation follows standardized benchmarks with broader category coverage, providing a more challenging and realistic setting.
We use Chamfer Distance (CD) to evaluate completion quality. CD measures the bidirectional nearest-neighbor distance between predicted and ground-truth point sets, reflecting both geometric coverage and reconstruction errors. As it does not require point ordering or explicit correspondences, CD is well-suited for unordered point clouds. Following prior work, we report both $l1$ and $l2$ CD on different datasets. We also adopt the F-score as an additional evaluation metric \cite{yu2021pointr}.

\subsection{Results on PCN dataset}
\label{sec:pcn_set}

\begin{table}
  \caption{ Quantitative results in terms of $l1$ Chamfer Distance $\times10^3$(CD) and F1-Score@$1\%$ (F1) on PCN dataset.}
  \label{tab:pcn}
  \centering
  \small
  \begin{tabular}{@{\hspace{5pt}}l@{\hspace{5pt}}|@{\hspace{5pt}}c@{\hspace{5pt}}c@{\hspace{5pt}}c@{\hspace{5pt}}c@{\hspace{5pt}}c@{\hspace{5pt}}c@{\hspace{5pt}}c@{\hspace{5pt}}c@{\hspace{5pt}}|@{\hspace{5pt}}c@{\hspace{5pt}}c@{\hspace{5pt}}}
    \toprule
    Methods & Airplane & Cabinet & Car & Chair & Lamp & Sofa & Table & Watercraft & CD ($\downarrow$) & F1 ($\uparrow$)\\
    \midrule
    PCN\cite{yuan2018pcn} & 5.50 & 22.7 & 10.63 & 8.70 & 11.00 & 11.34 & 11.68 & 8.59 & 9.64 & 0.695  \\
    AdaPoinTr\cite{yu2021pointr} & 3.68 & 8.82 & 7.47 & 6.85 & 5.47 & 8.35 & 5.80 & 5.76 & 6.53 & - \\
    SVDFormer\cite{zhu2023svdformer} & 3.62 & 8.79 & 7.46 & 6.91 & 5.33 & 8.49 & 5.90 & 5.83 & 6.54 & 0.841\\
    CRA-PCN\cite{rong2024cra} & 3.62 & 8.77 & 7.00 & 6.92 & 5.46 & 8.59 & 6.27 & 5.86 & 6.56 & 0.846\\
    T-CorresNet\cite{duan2024t} & 3.63 & 9.79 & 7.47 & 6.85 & 5.47 & 8.35 & 5.80 & 5.76 & 6.53 & 0.845\\
    DC-PCN\cite{wu2025dc} & 3.65 & 8.75 & 7.48 & 6.71 & 5.35 & 8.28 & 5.76 & 5.71 & 6.46 & 0.850\\
    SymmCompletion\cite{yan2025symmcompletion} & 3.56 & 8.51 & 7.34 & 6.56 & 5.09 & 8.40 & 5.71 & 5.52 & 6.33 & 0.852  \\
    \midrule
    AdaPoinTr + Ours & 3.56 & 8.80 & 7.43 & 6.87 & 5.24 & 8.42 & 5.91 & 5.67 & 6.48 & - \\
    Ours & \textbf{3.53} & \textbf{8.50} & \textbf{7.30} & \textbf{6.55} & \textbf{4.95} & \textbf{8.25} & \textbf{5.65} & \textbf{5.52} & \textbf{6.27} & \textbf{0.855}  \\
    \bottomrule
  \end{tabular}
\end{table}

\begin{figure}
  \centering
  \includegraphics[width=1\linewidth]{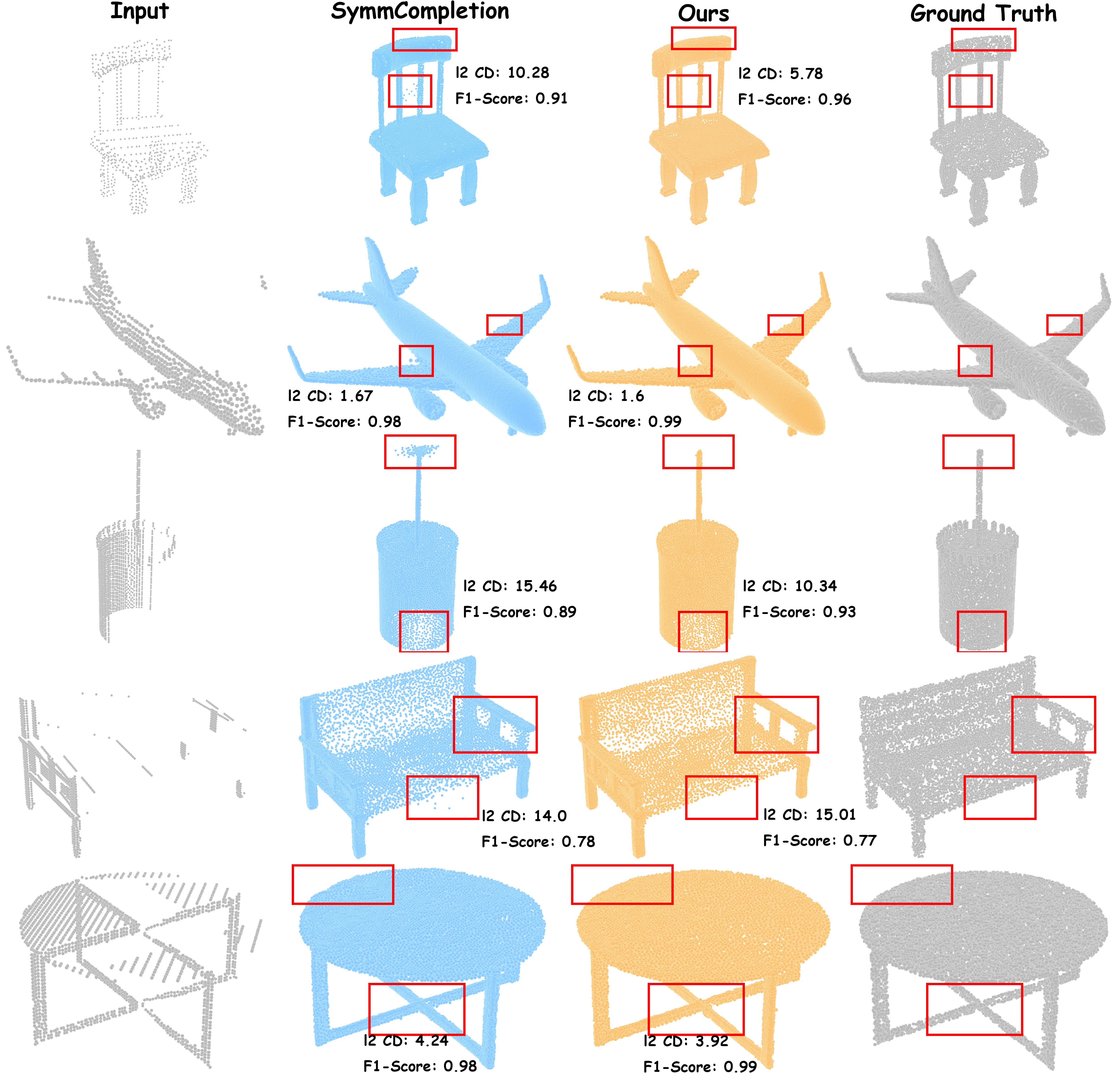}
  \caption{Qualitative comparison between our method and prior works. Our method not only generates high-quality, high-density outputs but also produces fewer local artifacts. We report the $l2$ Chamfer Distance $\times10^5$ (CD, $\downarrow$) and F1-Score@$1\%$ (F1, $\uparrow$) for generation results.}
  \label{fig:pcn_compare}
\end{figure}



We first report quantitative comparisons on the PCN dataset using $l1$ CD and F1-score. As shown in Tab. \ref{tab:pcn}, our method achieves the best overall $l1$ CD and F1-score on PCN. It also obtains the best or competitive category-level results across the eight categories, indicating improved global structure recovery and local detail reconstruction. We also evaluate AdaPoinTr enhanced with our method, which shows improvement over its original performance.
 
Fig. \ref{fig:pcn_compare} presents qualitative comparisons with SymmCompletion. Our method produces more consistent global structures and finer local details, including sharper chair-back slats and smoother surfaces, more accurate airplane wings, improved thin structures (e.g., wires), and better reconstruction of small components like handles and table legs. Overall, our results are closer to the ground truth with fewer artifacts.

\subsection{Results on ShapeNet-55/34}

\begin{table}
  \caption{Quantitative results in terms of $l2$ Chamfer Distance $\times10^3$(CD, $\downarrow$) on ShapeNet55/34 dataset for three difficulty levels.}
  \label{tab:shapenet}
  \centering
  \small
  \begin{tabular}{l|ccc|ccc|ccc}
    \toprule
    \multirow{2}{*}{Method} & \multicolumn{3}{c}{ShapeNet-55} & \multicolumn{3}{c}{ShapeNet-34} & \multicolumn{3}{c}{Unseen 21}\\
             & CD-S & CD-M & CD-H & CD-S & CD-M & CD-H & CD-S & CD-M & CD-H \\
    \midrule
    PCN\cite{yuan2018pcn} & 1.94 & 1.96 & 4.08 & 1.87 & 1.81 & 2.97 & 3.17 & 3.08 & 5.29 \\
    PoinTr\cite{yu2021pointr} & 0.58 & 0.88 & 1.79 & 0.76 & 1.05 & 1.88 & 1.04 & 1.67 & 3.44 \\
    AdaPoinTr\cite{yu2021pointr} & 0.49 & 0.69 & 1.24 & 0.48 & 0.63 & 1.07 & 0.61 & 0.96 & 2.11 \\
    SeedFormer\cite{zhou2022seedformer} & 0.50 & 0.88 & 1.79 & 0.48 & 0.70 & 1.30 & 0.61 & 1.07 & 2.35 \\
    SVDFormer\cite{zhu2023svdformer} & 0.48 & 0.70 & 1.30 & 0.46 & 0.64 & 1.13 & 0.61 & 1.05 & 2.19 \\
    CRA-PCN\cite{rong2024cra} & 0.48 & 0.71 & 1.37 & 0.45 & 0.65 & 1.18 & 0.55 & 0.97 & 2.19 \\
    SymmCompletion\cite{yan2025symmcompletion} & 0.36 & 0.55 & 1.12 & 0.33 & 0.48 & 1.00 & \textbf{0.39} & 0.70 & 1.83 \\
    \midrule
    Ours & \textbf{0.32} & \textbf{0.47} & \textbf{0.94} & \textbf{0.33} & \textbf{0.47} & \textbf{0.91} & 0.43 & \textbf{0.70} & \textbf{1.70} \\
    \bottomrule
  \end{tabular}
\end{table}

For the ShapeNet-55/34 dataset, we also perform a comprehensive quantitative evaluation against SOTA methods. This dataset contains three difficulty subsets.
We first train and validate on all 55 categories, then train on 34 categories, and test on both 34 seen categories and 21 unseen categories to evaluate our method's generalisation ability.

The experimental results are shown in Tab. 2, where CD-S, CD-M and CD-H represent the $l2$ CD at the three difficulty levels. Compared to existing methods, our method achieves strong performance across most settings, with clear improvements on ShapeNet-55 and strong results on both seen and unseen categories. On the more challenging Unseen-21 split, SUMI remains competitive and achieves the best result under the hardest missing setting, suggesting good robustness to unseen categories.




\subsection{Results on MVP}

\begin{table}
  \caption{Quantitative results in terms of $l2$ Chamfer Distance $\times10^4$ (CD) and F1-Score@$1\%$ (F1) on MVP dataset for four densities.}
  \label{tab:mvp}
  \centering
  \small
  \begin{tabular}{l|cc|cc|cc|cc}
    \toprule
    \multirow{2}{*}{Method}  & \multicolumn{2}{c}{2048} & \multicolumn{2}{c}{4096} & \multicolumn{2}{c}{8192} & \multicolumn{2}{c}{16384}\\
             & CD ($\downarrow$) & F1 ($\uparrow$)& CD ($\downarrow$) & F1 ($\uparrow$)& CD ($\downarrow$) & F1($\uparrow$)& CD ($\downarrow$) & F1($\uparrow$)\\
    \midrule
    PCN\cite{yuan2018pcn} & 9.77 & 0.32 & 7.96 & 0.46 & 6.99 & 0.56&6.02&0.64\\
    SnowflakeNet \cite{xiang2021snowflakenet} & 5.71 & 0.50 & 4.45 & 0.65 & 3.48 & 0.74 & 2.69 & 0.79 \\
    PDR \cite{lyu2021conditional} & 5.66 & 0.49 & 4.26 & 0.65 & 3.35 & 0.75 & 2.61 & 0.82 \\
    AEDNet \cite{fu2024aednet}& 5.12 & 0.52 & 3.75 & 0.66 & 3.21 & 0.76 & 2.24 & 0.83 \\
    SymmCompletion\cite{yan2025symmcompletion} & 4.89 & \textbf{0.54} & 3.65 & 0.69 & 2.70 & 0.78&2.14&0.85\\
    \midrule
    Ours & \textbf{4.85} & 0.53 & \textbf{3.46} & \textbf{0.69} & \textbf{2.68} & \textbf{0.80}&\textbf{2.09}&\textbf{0.86}\\
    \bottomrule
  \end{tabular}
\end{table}

We next validate the completion performance on the MVP dataset. Tab. \ref{tab:mvp} shows the performance compared with prior works at 4 different densities. In terms of CD, our method achieves the best results for all 4 densities. For F1-score, our method matches the best result at 4096 points and achieves the best performance at 8192 and 16384 points.

\begin{figure}[h]
  \centering
  \begin{minipage}[b]{0.25\columnwidth}
    \centering
    \includegraphics[width=\linewidth]{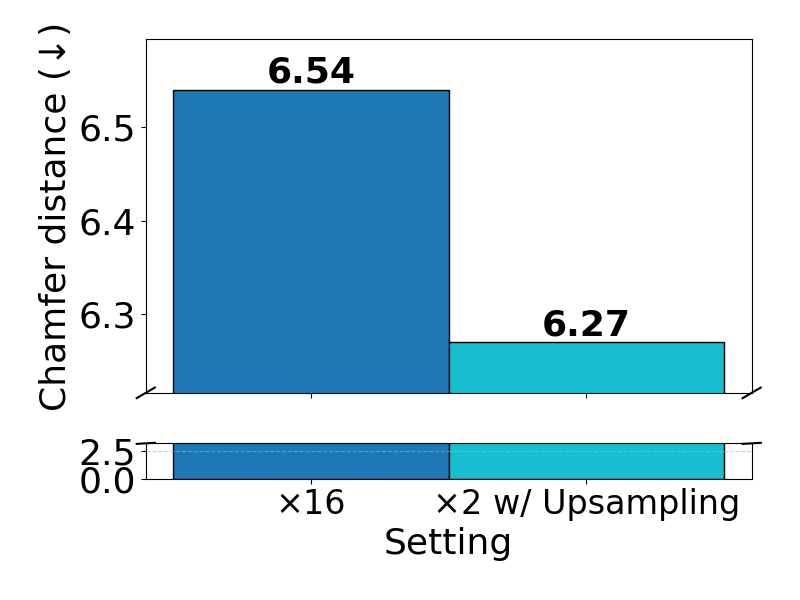}
    
    \caption{Ablation study on output density for SUMI Module.}
    \label{fig:abl_1}
  \end{minipage}
  \hfill
  \begin{minipage}[b]{0.25\columnwidth}
    \centering
    \includegraphics[width=\linewidth]{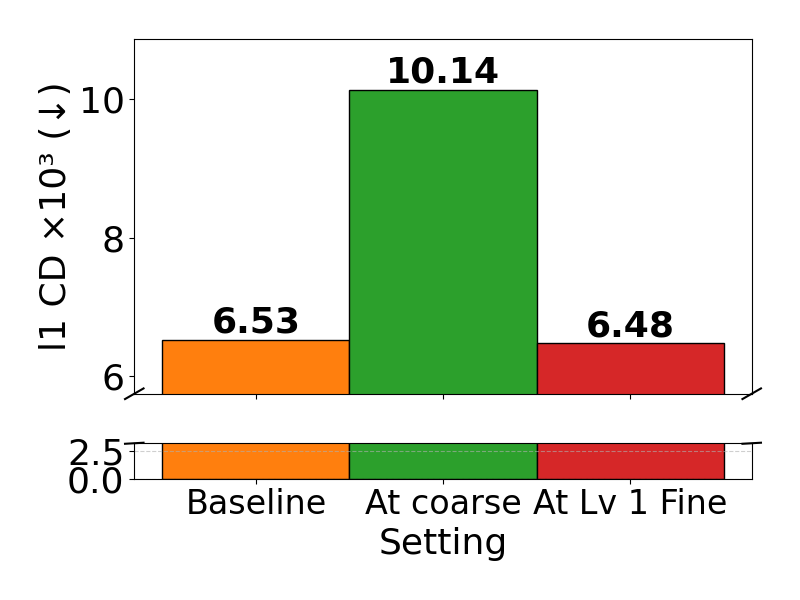}
    
    \caption{Ablation study on location setting for SUMI Module.}
    \label{fig:abl_2}
  \end{minipage}
  \hfill
  \begin{minipage}[b]{0.25\columnwidth}
    \centering
    \includegraphics[width=\linewidth]{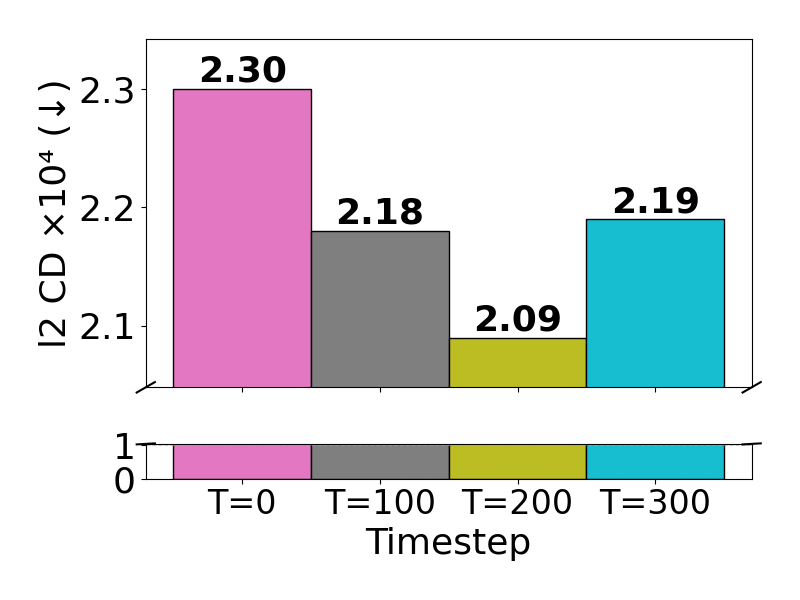}
    
    \caption{Ablation study on different timesteps for SUMI Module.}
    \label{fig:abl_3}
  \end{minipage}
  \label{fig:two_side}
\end{figure}
\subsection{Ablation Study}
\label{sec:abl}
We conduct ablation studies on PCN to evaluate the impact of the SUMI Module and diffusion process. Removing the upsampling module and directly generating 16384 points leads to degraded performance (Fig. \ref{fig:abl_1}), suggesting that large-scale upsampling hinders fine detail reconstruction. In contrast, generating intermediate 2048-point outputs with SUMI and applying subsequent upsampling preserves more geometric details. 

We next examine the insertion stage of SUMI within AdaPoinTr on PCN dataset. As shown in Fig. \ref{fig:abl_2}, applying SUMI at the coarse stage significantly degrades performance due to early noise interference, while introducing it at the fine stage yields stable improvements, indicating its effectiveness for local refinement. 

We further investigate the effect of diffusion timesteps $T$ on the MVP dataset. As shown in Fig.~\ref{fig:abl_3}, we evaluate four settings from $T=0$ to $T=300$. The $T=0$ setting keeps the SUMI refinement architecture but disables iterative reverse denoising, serving as a same-architecture single-step refinement baseline. Performance consistently improves as $T$ increases and reaches the best result at $T=200$, indicating that the gain comes not only from the additional refinement module but also from the diffusion-based iterative denoising process. Increasing the timestep further to $T=300$ leads to performance degradation, suggesting that $T=200$ provides a better balance between reconstruction quality and inference efficiency.

\section{Limitation}
A limitation of SUMI is that iterative denoising introduces additional inference cost compared with purely feed-forward completion models. To mitigate this, we apply SUMI only at the first fine-generation stage and use a lightweight upsampling module for the final high-density output. Future work may explore accelerated sampling or distillation to further improve efficiency.

\section{Conclusion}
In this paper, we present a diffusion-enhanced coarse-to-fine framework for PCC. Extensive experiments across multiple benchmarks demonstrate the effectiveness of our approach, achieving competitive or state-of-the-art performance in point cloud completion. Furthermore, SUMI can be integrated into existing coarse-to-fine architectures as a refinement module, showing its potential to enhance future PCC methods.

\bibliographystyle{unsrtnat}
\bibliography{ref}

\newpage

\section*{Supplementary Material}
\setcounter{section}{0}
\setcounter{table}{0}
\renewcommand{\thetable}{A\arabic{table}}
\setcounter{figure}{0}
\section{Computational Resources and Efficiency}

\subsection{Hardware and Software Environment}

All experiments were conducted on a single NVIDIA RTX 4090 GPU. The detailed hardware and software environment is summarized in Tab.~\ref{tab:hardware}.

\begin{table}[h]
\centering
\caption{Hardware and software environment used in our experiments.}
\label{tab:hardware}
\begin{tabular}{ll}
\toprule
Item & Configuration \\
\midrule
GPU & NVIDIA RTX 4090, 24GB memory \\
CPU & Intel(R) Core(TM) i7-13700K \\
System memory & 64GB \\
Operating system & Ubuntu 22.04.4 LTS \\
CUDA version & 12.2 \\
Python version & 3.7 \\
PyTorch version & 1.13.1 \\
Training precision & FP32 \\
\bottomrule
\end{tabular}
\end{table}

\subsection{Training Cost}

We report the per-epoch training time and evaluation time for SUMI on different datasets in Tab.~\ref{tab:training_cost}. The reported training time includes the full end-to-end optimization of the coarse prediction module, the SUMI refinement module, and the final upsampling module unless otherwise specified. 

\begin{table}[h]
\centering
\caption{Training cost of SUMI on different datasets.}
\label{tab:training_cost}
\begin{tabular}{lccccc}
\toprule
\multirow{2}{*}{Dataset} & \multirow{2}{*}{GPU} & \multirow{2}{*}{Batch size} & \multirow{2}{*}{Epochs} & Training  & Evaluation \\
& & & & Mins / Epoch & Mins / Epoch\\
\midrule
PCN & RTX 4090 & 16 & 350 & 10 & 20\\
ShapeNet-55 & RTX 4090 & 16 & 350 & 20 & 140\\
ShapeNet-34 & RTX 4090 & 16 & 350 & 20 & 50\\
MVP & RTX 4090 & 16 & 40 & 22 & 550\\
\bottomrule
\end{tabular}
\end{table}


\section{Diffusion Formulation}

SUMI follows the standard DDPM forward diffusion process~\cite{ho2020denoising}. Given a clean complete point cloud $x_0$, the forward process gradually perturbs it into a noisy point cloud $x_t$:
\[
q(x_t \mid x_{t-1}) =
\mathcal{N}\left(x_t; \sqrt{1-\beta_t}x_{t-1}, \beta_t I\right),
\]
where $\beta_t$ is the noise variance at timestep $t$. Equivalently, $x_t$ can be sampled directly from $x_0$ as:
\[
q(x_t \mid x_0) =
\mathcal{N}\left(x_t; \sqrt{\bar{\alpha}_t}x_0, (1-\bar{\alpha}_t)I\right),
\]
where $\alpha_t = 1-\beta_t$ and $\bar{\alpha}_t=\prod_{s=1}^{t}\alpha_s$.

Unlike standard DDPMs that commonly train the network to predict the Gaussian noise, SUMI predicts the clean point cloud representation from the noisy observation. Specifically, given the noisy point cloud $x_t$, the coarse prediction $P_c$, keypoint features $F_k$, and symmetry features $F_s$, SUMI predicts a fine-level point cloud $x_f$. We supervise this prediction using Chamfer Distance:
\[
\mathcal{L}_{diff} = \mathcal{L}_{CD}(x_f, x_0).
\]
This objective is better aligned with point cloud completion, where the goal is to reconstruct missing geometry conditioned on the partial observation and coarse structural features.

\section{Notation}

For clarity, we summarize the main notation used in the paper in Tab.~\ref{tab:notation}. 
All symbols are also defined at their first occurrence in the main paper.

\begin{table}[h]
\centering
\caption{Summary of notation used in SUMI.}
\label{tab:notation}
\begin{tabular}{ll}
\toprule
Symbol & Description \\
\midrule
$x_0$ & Ground-truth complete point cloud \\
$x_t$ & Noisy point cloud at diffusion timestep $t$ \\
$T$ & Total number of diffusion timesteps \\
$t$ & Current diffusion timestep \\
$P_c$ & Coarse prediction generated by the coarse module \\
$x_f$ & Fine-level prediction generated by the SUMI module \\
$P_f$ & Fine output from the SUMI module \\
$P''_f$ & Final high-density output after upsampling \\
$F_c$ & Feature representation encoded from the coarse prediction $P_c$ \\
$F_x^t$ & Feature representation encoded from the noisy point cloud $x_t$ \\
$F_k$ & Keypoint features extracted from the coarse-generation module \\
$F_s$ & Symmetry features extracted from the coarse-generation module \\
$F_{\text{fusion}}$ & Fused feature after noise-conditioned cross-attention \\
$F_{kf}$ & Feature obtained by fusing $F_{\text{fusion}}$ with keypoint features \\
$F_{sf}$ & Feature obtained by fusing $F_{\text{fusion}}$ with symmetry features \\
$\mathcal{C}(\cdot)$ & Cross-attention operation \\
$\mathcal{S}(\cdot)$ & Self-attention operation \\
$\mathcal{L}_{CD}$ & Chamfer Distance loss \\
$\mathcal{L}_{diff}$ & Diffusion reconstruction loss \\
$\mathcal{L}$ & Overall training objective \\
\bottomrule
\end{tabular}
\end{table}







\end{document}